\documentclass[runningheads]{llncs}

\usepackage{accv}

\usepackage{accvabbrv}

\usepackage{graphicx}
\usepackage{booktabs}
\usepackage{tabularx}
\usepackage{float}
\usepackage{xcolor}
\usepackage{hyperref}

\usepackage[accsupp]{axessibility}  

\usepackage{hyperref}

\usepackage{orcidlink}
\begin{document}

\title{KhmerST: A Low-Resource Khmer Scene Text Detection and Recognition Benchmark}

\titlerunning{KhmerST Dataset for Low-Resource STDR}

\author{
    Vannkinh Nom\inst{1,2}\orcidlink{0009-0004-4684-0416} \and
    Souhail Bakkali\inst{1}\orcidlink{0000-0001-9383-3842} \and
    Muhammad Muzzamil Luqman\inst{1}\orcidlink{0000-0002-9658-0833} \and
    Micka\"{e}l Coustaty\inst{1}\orcidlink{0000-0002-0123-439X} \and 
    Jean-Marc Ogier\inst{1}\orcidlink{0000-0002-5666-475X}
}

\authorrunning{Nom et al.}

\institute{La Rochelle University, Laboratoire Informatique Image Interaction (L3i) \and
Cambodia Academy of Digital Technology\\
\email{\{vannkinh.nom, souhail.bakkali, muhammad\_muzzamil.luqman, mickael.coustaty, jean-marc.ogier\}@univ-lr.fr}}

\maketitle

\begin{abstract}
Developing effective scene text detection and recognition models hinges on extensive training data, which can be both laborious and costly to obtain, especially for low-resourced languages. Conventional methods tailored for Latin characters often falter with non-Latin scripts due to challenges like character stacking, diacritics, and variable character widths without clear word boundaries. In this paper, we introduce the first Khmer scene-text dataset, featuring 1,544 expert-annotated images, including 997 indoor and 547 outdoor scenes. This diverse dataset includes flat text, raised text, poorly illuminated text, distant and partially obscured text. Annotations provide line-level text and polygonal bounding box coordinates for each scene. 
The benchmark includes baseline models for scene-text detection and recognition tasks, providing a robust starting point for future research endeavors.
The KhmerST dataset is publicly accessible\setcounter{footnote}{0}
\hspace{-2pt}\footnote{\url{https://gitlab.com/vannkinhnom123/khmerst}}.
\end{abstract}

\section{Introduction}
\label{sec:introduction}
\begin{figure} [tb]
  \centering
  \begin{subfigure}{\textwidth}  
    \includegraphics[width=\linewidth]{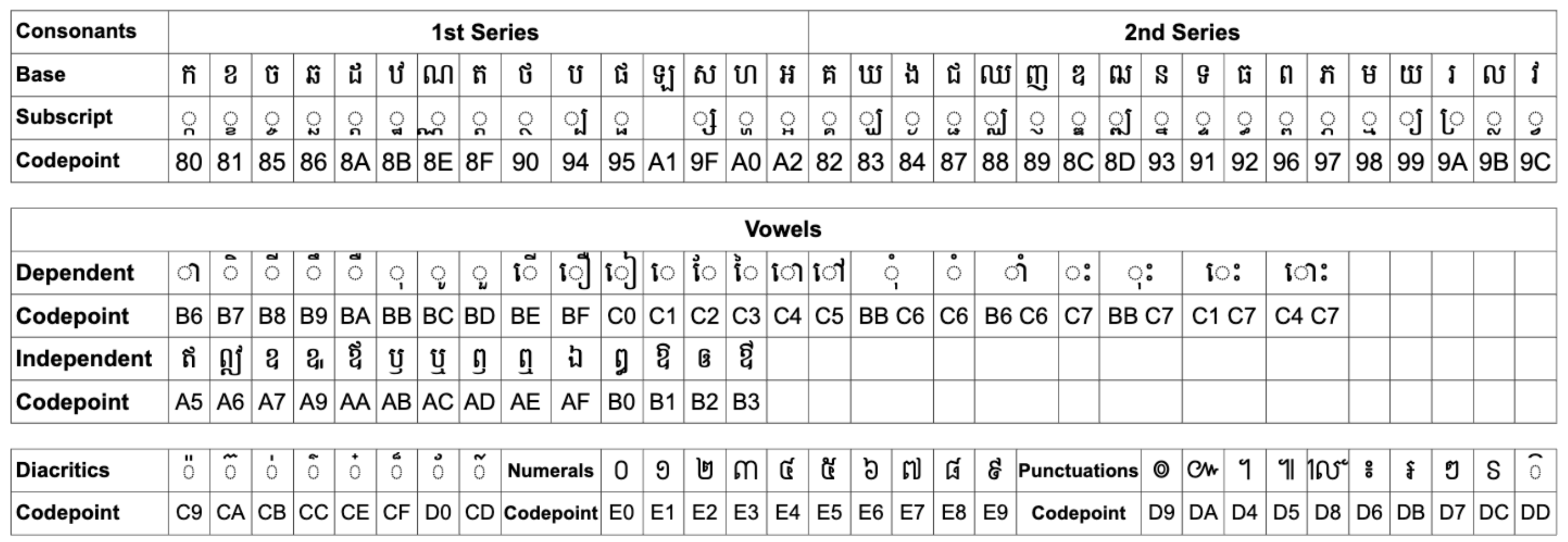}
    \caption{The Unicode point representation of the entire Khmer alphabet, including consonants, vowels, diacritics, and punctuation. Adapted from~\cite{buoy2023toward}.}
    \label{fig:unicode}
  \end{subfigure}
  
  \begin{subfigure}{\textwidth}  
    \includegraphics[width=\linewidth]{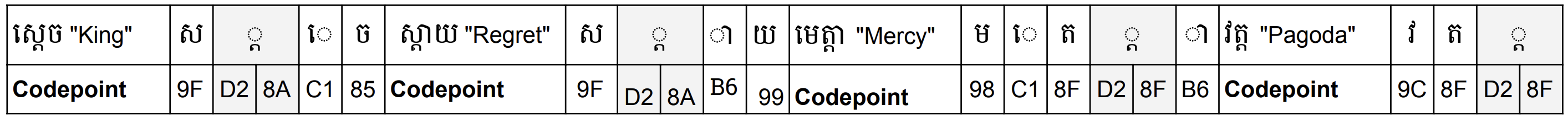}
    \caption{The Unicode point representation of word forms using "coeng (U17D2)". For the first two words, they use the subscript DA; for the last two words, they use the subscript TA, but both appear the same, while the Unicode points are different.}
    \label{fig:word-form}
  \end{subfigure}
  \caption{Illustration of the complexity in how Unicode points encode the Khmer alphabet and words.}
  \label{fig:1}
\end{figure}
\begin{figure}[tb]
  \centering
  \begin{subfigure}{\textwidth}  
    \includegraphics[width=0.97\linewidth]{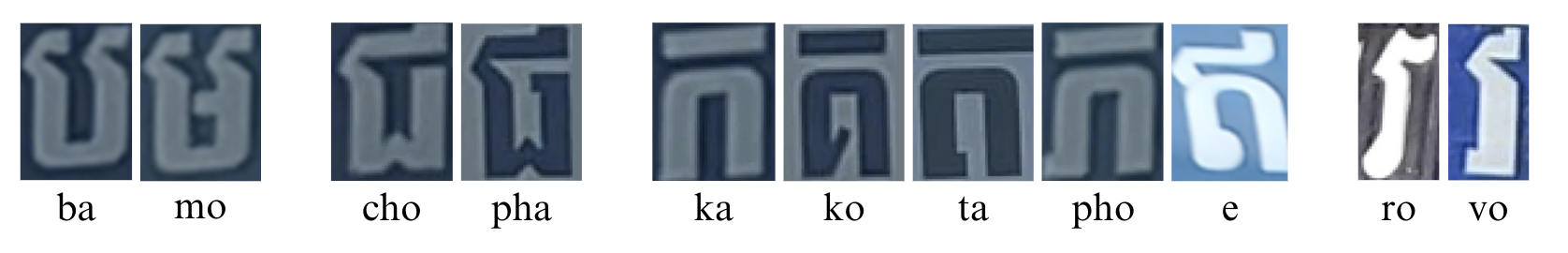}
    \caption{Different characters having similar appearances}
    \label{fig:char-1a}
  \end{subfigure}
  \begin{subfigure}{\textwidth}  
    \includegraphics[width=0.97\linewidth]{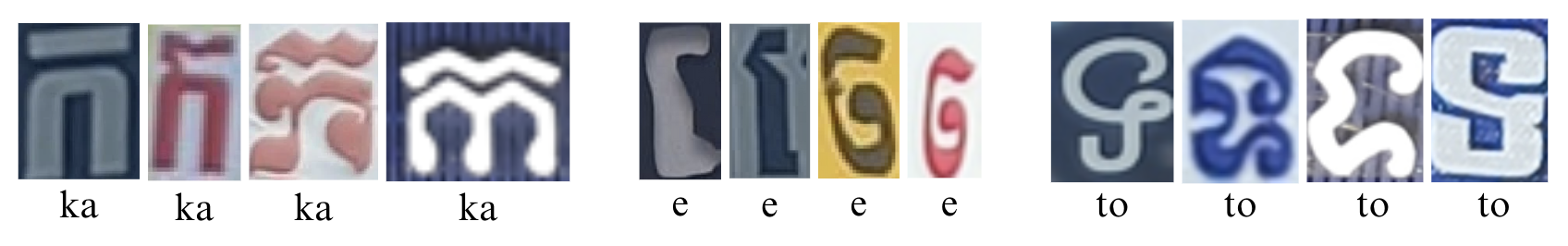}
    \caption{Same characters having different appearances}
    \label{fig:char-1b}
  \end{subfigure}
  \caption{The characters in (a) and (b) show the low inter-class variance versus high intra-class, which highlights the difficulty in Khmer writing. Group (a) demonstrates different characters appearing in similar shapes, while group (b) shows characters that appear in different shapes, although  having the same characteristics.}
  \label{fig:2}
\end{figure}
\begin{figure}[tb] 
    \centering
    \includegraphics[width=\linewidth]{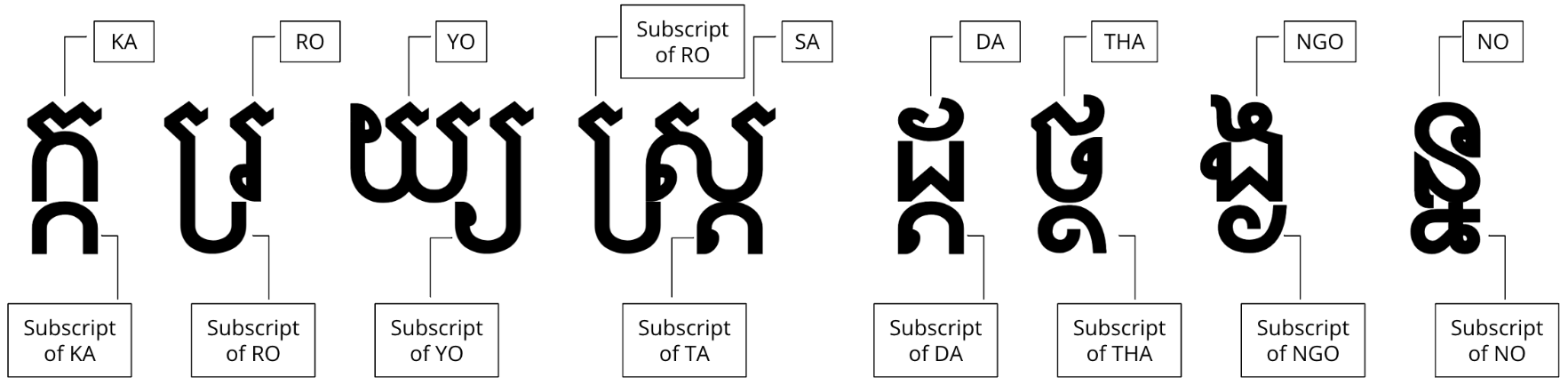} 
    \caption{Examples of multiple consonant clusters in Khmer, showing the different shapes of the consonants when they function as main or sub-consonants. Some consonants have only one sub-consonant, while others have more to form a word.}
    \label{fig:3}
\end{figure}

Automatic scene-text detection and recognition (STDR) in natural scenes are critical tasks in computer vision, with applications ranging from word spotting~\cite{gomez2017textproposals}, text-line detection~\cite{ yuan2019large}, character-level and/or word-level recognition~\cite{buoy2023toward, yuan2019large}. This problem has been extensively studied and is divided into two scenarios of varying difficulties: text detection and recognition in natural scenes, which presents significant challenges. In natural scene images, characters can vary greatly in appearance due to differences in style, font, resolution, and illumination. Additionally, text may be partially obscured, distorted, or set against complex backgrounds, complicating detection and recognition. The situation is further complicated by high intra-class variability (\ie differences within the same character) and low inter-class variability (\ie differences between different characters). Existing STDR methods, often designed for Latin scripts, struggle with non-Latin scripts like Khmer, which feature complex characteristics such as character stacking, diacritics, ligatures, non-uniform character widths, and the absence of explicit word boundaries. These complexities, combined with the low-resource nature of many non-Latin languages, necessitate advanced solutions.

In regard of non-Latin scripts like Khmer, the current state-of-the-art approaches in STDR struggle to fully address the aforementioned challenges. One of the main reasons is the complexity of Khmer characters, which consist of 33 consonants, 16 dependent vowels, and 14 independent vowels and 13 diacritics~\cite{horton2017spoof}. Khmer symbols are encoded using Unicode (U1780-U17FF), but the order of codes does not always follow the left-to-right writing direction. The relationship between symbols and codes is complex, with some symbols represented by multiple codes and some codes representing combinations of symbols. Since the subscripts don't have their own code, Khmer words are formed using a sequence of two codes: the "coeng (U17D2)" code, which is combined with the main consonant. Distinct from other languages, recognizing a Khmer word requires considering the entire word's writing, emphasizing on the importance of spatial information for accurate recognition. As specified by the Guinness World Records, Khmer script has the longest alphabet, consisting of 74 distinct characters~\cite{valy2019text}. ~\cref{fig:1} shows about the complexity of how Unicode point representation of Khmer alphabets and words. 
Depending on the fonts used, certain pairs of characters can be highly ambiguous, differing only by a single stroke. In extreme cases, some subscript forms are almost indistinguishable. Additionally, while some characters are formed as a single continuous glyph, others comprise multiple disconnected glyphs, each representing a distinct character. Furthermore, certain groups of characters appear similar in shape but are distinct, while others appear different due to the writing style and the different font types, as demonstrated in~\cref{fig:2}.
Unlike Latin scripts, Khmer forms words differently. On the one hand, Khmer script has a distinctive feature: consonants can have different shapes based on their spatial position related to the word, including main and sub-consonants, can feature the merging of two or more consonants into different shapes, known as low-consonants or subscripts, which are located beneath the main consonant as illustrated in~\cref{fig:3}. This combination helps to create the desired consonant sound~\cite{valy2017new}. Vowels, on the other hand, can appear everywhere \ie before, after, above, and below consonants. In some special cases, vowels can combine with each other to create new vowels. Additionally, the absence of word separation in Khmer writing makes the detection task even more challenging compared to languages that have distinct word boundaries. Moreover, challenges related to low contrast against complex backgrounds, varying lighting conditions, and non-standard text orientations contribute to the difficulty in achieving higher accuracy and reliability. These factors make Khmer text in scene images more difficult to detect and to accurately recognize.


In this work, we present the first dataset of Khmer text in natural scene images, named KhmerST (\textbf{Khmer} \textbf{S}cene-\textbf{T}ext). The dataset contains 1,544 expert-annotated images, including 997 indoor and 547 outdoor scenes, making it a diverse and complex collection. The challenges within this dataset include planar text, raised text, poorly illuminated text, distant text, and partially occluded text. Each scene image is annotated with line-level associated text and polygonal bounding box coordinates. We also benchmark the dataset using several state-of-the-art approaches for text detection and recognition. By introducing KhmerST, we aim to provide a valuable resource for future research on detection and recognition tasks, as well as text segmentation and word spotting of Khmer in natural scene images.
The novelty of the KhmerST dataset lies in its comprehensive approach to capturing the uniqueness of the Khmer script in diverse real-world scenarios. This focus supports the development of more effective and inclusive computer vision technologies, filling a significant gap in resources tailored to Southeast Asian scripts, especially Khmer. Unlike most existing datasets that primarily focus on Latin, Chinese, or Arabic scripts, KhmerST provides an essential resource for creating solutions finely tuned to the needs of the Cambodian population. These applications include digital archiving of documents, automated translation services, and enhanced accessibility features for technology applications in Khmer. By capturing a variety of text appearances and settings not present in synthetic datasets, KhmerST serves as a more comprehensive and challenging resource for OCR development, enabling robust model training to handle a wide range of practical situations.

Therefore, this work makes two key contributions. First, we introduce KhmerST, a novel low-resource benchmark specifically designed for Khmer scene-text. Second, we demonstrate that the performance of current state-of-the-art models on KhmerST significantly lags behind their performance on Latin script benchmarks. This discrepancy underscores the necessity for more holistic and efficient modeling approaches tailored to the complexities of the Khmer script.

The paper is structured as follows. \cref{sec:related-work} reviews the related work. \cref{sec:Khmer-STD} introduces the KhmerST dataset and its characteristics, the detailed process of collecting the data, the annotation, while \cref{sec:benchmark} presents the tasks, the baselines and the evaluation metrics, followed with a discussion. Finally, \cref{sec:conclusion} offers a conclusion.


\section{Related work}
\label{sec:related-work}

\subsection{Datasets of Text in Natural Scene Images}

As STDR is a popular research domain in the computer vision community, numerous dataset benchmarks are available in various languages. These datasets are classified into two main categories: real-world text and synthetic text. For real-world text datasets~\cite{lucas2005icdar, mishra2012scene, veit2016coco}, they offer multilingual text images captured in natural scenes that contain text from urban environments and serve as a valuable benchmark for evaluating recognition algorithms. On the other hand,~\cite{de2009character, jaderberg2014synthetic} propose the synthetic text dataset that was generated using computer vision techniques and produced a huge amount of labeled data. Besides, Google Street View images, which also count as a dataset for STDR belongs to the real-world text category. These datasets are particularly beneficial to facilitate the development of robust STDR systems. The Street View House Number (SVHN) dataset, presented by Netzer \etal~\cite{netzer2011reading}, contains images of house numbers collected from Google Street View, providing a challenge for digit recognition due to the various fonts, sizes, and backgrounds of the images. The FSNS dataset, introduced by Smith \etal~\cite{smith2016end}, consists of more than a million images from Google Street View in France. These images contain street name signs, which help address street name extraction problems. The CTW dataset, presented by Yuan \etal~\cite{yuan2019large}, contains a large collection of Chinese text with over 30,000 street view images. This dataset is an important resource for evaluating and developing scene text recognition systems for Chinese. The KAIST dataset, introduced by Jung \etal~\cite{jung2011touch}, contains Korean text information and image ground truth, encompassing a wide range of scene images that present text in different formats and contexts. For the Khmer language, there is a synthetic dataset proposed by Bouy \etal~\cite{buoy2023toward}. The dataset contains a large number of images of Khmer text. However, a comprehensive Khmer scene text dataset does not exist. The absence of such a dataset presents significant challenges for developing and evaluating scene text recognition systems for the Khmer language. Furthermore, we should have real-world data to better evaluate the capacity of existing systems on the Khmer script. Developing this dataset would fill a critical gap and advance optical character recognition (OCR) technology for future applications in Khmer.

\subsection{Scene Text Detection and Recognition}
The traditional approaches in scene-text detection focused on hand-crafted low-level features to differentiate text and non-text components in scene images. The sliding window  (SW) method detects text information by moving the sub-window through all locations of the image; it utilizes the pre-train classifier to ensure whether text exists within the sub-window, as described by He \etal~\cite{he2016text}. Wang \etal~\cite{wang2012end} conducted a convolutional neural network (CNN) with the SW method to find the position of text on the images. The connected component-based methods, as described by Zhu \etal~\cite{zhu2016scene}, are designed to extract the components from the image and filter out non-text candidates using manually designed rules or automatically trained classifiers. Recently, there has been much attention on deep learning in semantic segmentation and general object detection. As a result, similar techniques are increasingly being utilized in text detection. Qin and Manduchi~\cite{qin2017cascaded} developed a text detection method using a cascade of two convolutional neural networks (CNNs). Firstly, text regions of interest are identified by a fully convolutional network (FCN) and resized to a square shape with a fixed size. Redmon \etal~\cite{redmon2016you} proposed YOLO (You Only Look Once), which is an object detection model that can predict the object and its location at a glance. It considers the entire image during training and test time, implicitly encoding contextual information about classes and their appearance. In our case, for object detection, we propose to use different versions of YOLO and consider the text as an object. 

As for the scene text recognition task, many approaches were considered in the literature such as character classification-based methods, word classification-based methods, sequence-based methods, end-to-end text spotting, \etc. Lee and Osindero~\cite{lee2016recursive} proposed recursive recurrent neural networks (RNNs) enhanced with an attention model for text recognition. Jaderberg \etal~\cite{jaderberg2014synthetic} experimented with a CNN framework to train the synthetic data with handcrafted labeling and receive a good performance for word recognition. Li \etal~\cite{li2023trocr} introduced transformer-based optical character recognition (TrOCR), an end-to-end text recognition approach with an image transformer and text transformer pre-train model. The TrOCR model can be used with large-scale synthetic data, printed text, and handwritten and scene text. 



\section{The KhmerST Dataset}
\label{sec:Khmer-STD}
The KhmerST dataset is a new collection specifically designed to advance computer vision research focused on the Khmer script. This dataset comprises numerous images captured from various public places in Cambodia, such as streets, signboards, supermarkets, and commercial establishments, all featuring text written in Khmer. To our knowledge, it is the first scene-text dataset for the Khmer language, making it a unique contribution compared to existing benchmark datasets that include Khmer printed text, as in~\cite{sok2014support, buoy2022khmer}, the historical handwritten Sleuk-Rith dataset\cite{valy2020data}, scanned books~\cite{fujii2017sequence}, synthetic documents, synthetic scene text, KHOB, and ID cards proposed by~\cite{buoy2023toward}.The KhmerST dataset is crucial because it provides real-world scenarios, illustrating how the language is used in everyday contexts, essential for developing robust and accurate text recognition models. It addresses challenges such as varying lighting conditions, diverse font styles, and background noise typical in natural scene images. Unlike other datasets, the KhmerST dataset captures real-world environments, which is essential for training models to handle a wide range of practical situations. Additionally, it includes a variety of text appearances and settings not present in synthetic datasets, making it a more comprehensive and challenging resource for OCR development.

The novelty of the KhmerST dataset lies in its comprehensive approach to capturing the uniqueness of the Khmer script in diverse real-world scenarios. This focus supports the development of more effective and inclusive computer vision technologies, filling a significant gap in resources tailored to Southeast Asian scripts, especially Khmer. Unlike most existing datasets that primarily focus on Latin, Chinese, or Arabic scripts, KhmerST provides an essential resource for creating solutions finely tuned to the needs of the Cambodian population. These applications include digital archiving of documents, automated translation services, and enhanced accessibility features for technology applications in Khmer. By capturing a variety of text appearances and settings not present in synthetic datasets, KhmerST serves as a more comprehensive and challenging resource for OCR development, enabling robust model training to handle a wide range of practical situations.

\subsection{Image Selection}
To create the KhmerST dataset, we embarked on a data collection process across Cambodia, amassing a total of 1,544 images. These images were captured from a variety of locations to ensure a broad representation of settings. We utilized four different smartphone models for this purpose: Samsung Galaxy A32, iPhone 8 Plus, iPhone 13 Pro Max, and iPhone 14 Pro Max. This diversity in devices helped to capture images under different lighting conditions and camera capabilities, enhancing the dataset's robustness. 

\begin{figure}[tb] 
    \centering
    \includegraphics[width=\linewidth]{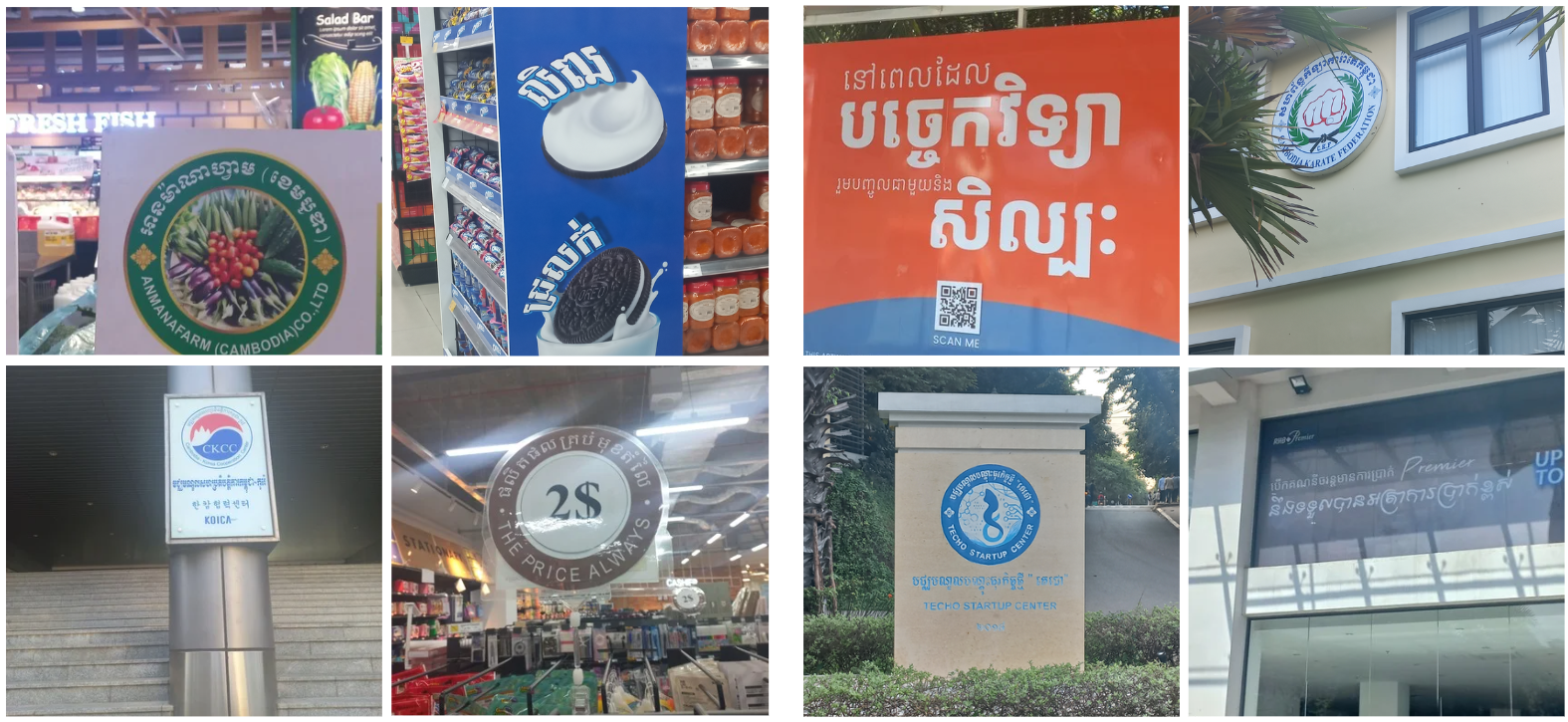}
    \caption{Examples of the KhmerST dataset: The Khmer script is present in both indoor and outdoor images. The dataset showcases the diversity of font sizes, styles, and the various ways the script appears, such as straight, rotated and curved text.}
    \label{fig:4}
\end{figure}

The KhmerST dataset is divided into two main categories: indoor and outdoor images. Indoor images feature text from commercial environments like supermarkets, while outdoor images include text from streets, signboards, and public buildings. \cref{fig:4} illustrates examples from both categories. The variety in font styles and the different ways the script appears (\eg straight, rotated, and curved text) ensures that the dataset can be used to develop robust text detection and recognition models capable of handling various real-life conditions.

\subsection{Dataset Annotation}
The VGG Image Annotator (VIA) was utilized for the annotation process. It is a powerful tool designed for marking up images with annotations. This annotator allows to define regions within each image using polygon coordinates, effectively delineating complex shapes by specifying vertices on the $x$ and $y$ axes. These annotations are crucial for precise object detection and region-specific analysis.

\begin{figure}[tb] 
    \centering
    \includegraphics[width=\linewidth]{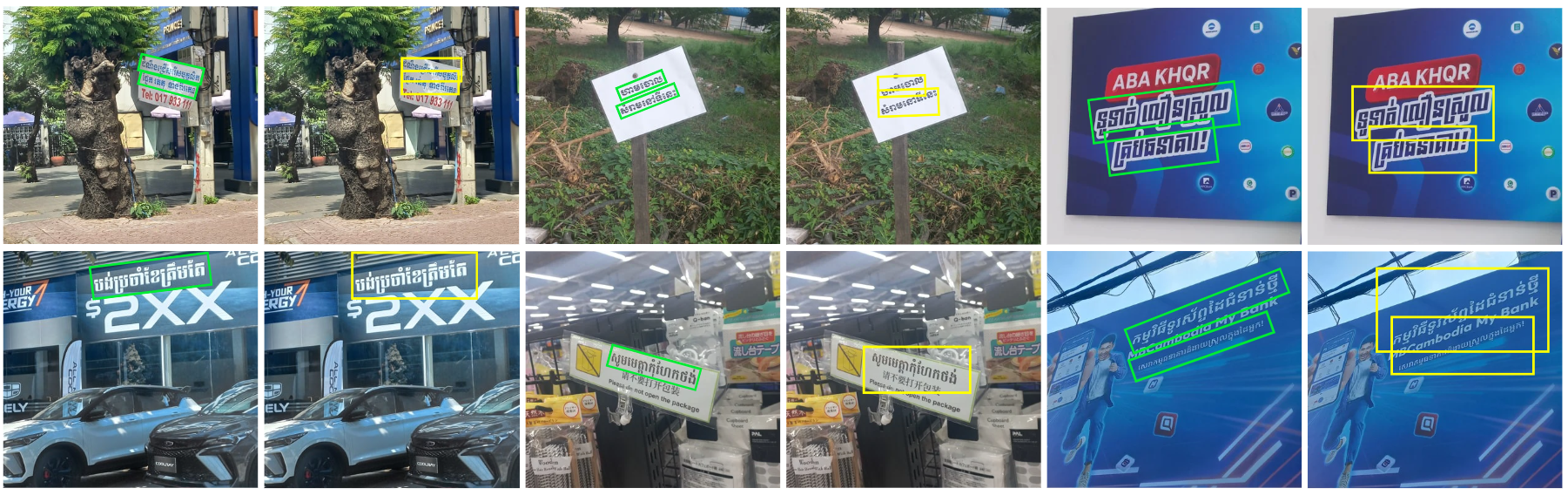}
    \caption{Examples of bounding boxes on text areas. The green bounding boxes represent polygon coordinates, while the yellow ones represent rectangular coordinates.}
    \label{fig:5}
\end{figure}

The data for each image, including its annotations, is structured in JSON format, offering a clear, hierarchical representation of attributes. In our JSON structure, each polygon's coordinates are represented by arrays of x and y points such as "all\_points\_x": $[x1, x2, x3, x4]$ and "all\_points\_y": $[y1, y2, y3, y4]$. This dataset format uses polygons linked to line-level text to describe text areas in images rather than rectangular coordinates, because polygons can adjust to the way the text appears (\eg rotated text). This method accommodates text rotations and contours, improving recognition accuracy. The JSON entries also include essential metadata, such as image filenames and sizes, to enhance the dataset's utility for training and evaluating machine learning models, particularly for Khmer script recognition. \cref{fig:5} demonstrates the advantage of using polygons over rectangles for capturing text areas.

\subsection{Dataset Splits}
\label{sec:splitting-data}
We divided our KhmerST dataset into training and test sets for the text detection task by allocating 80\% of the 1,544 images to the training set and 20\% to the test set. This split results in 1,236 training images and 308 testing images, combining both outdoor and indoor categories to enhance diversity and challenge. The KhmerST dataset is paired with images and JSON files containing detailed text information. For the text recognition task, we cropped the text regions from all images, yielding a total of 3,463 cropped images. We applied the same 80/20 split, resulting in 2,851 training images and 712 test images. This systematic division ensures a robust evaluation framework for both detection and recognition tasks.


\section{Benchmark Tasks and Evaluation Metrics}
\label{sec:benchmark}
In this section, we outline the performance of the baseline models along with the evaluation metrics on the proposed KhmerST dataset. The experiments were conducted on an NVIDIA RTX A6000 GPU and 252 GB of RAM. The process includes two modules: Text Detection and Text Recognition.

\subsection{Scene Text-Line Detection}
Given an input image, our challenge is to detect the bounding box regions of Khmer text at line-level. The ground truth for each text line's bounding box region is provided. To address this challenge, we experimented with four different YOLO models. The decision to utilize YOLO models for text-line detection is based on their unique strengths, which align well with the complexities of text recognition in varied and dynamic environments.

\noindent\textbf{Evaluation Metrics.} For the detection task, the Intersection over Union (IOU) metric is used to measure how well the predicted bounding boxes overlap with the actual ones. An IOU score ranges from 0 to 1, where 1 indicates a perfect match. Higher IOU score reflect more accurate object detection, while a lower score suggest inaccuracies in the predictions. To determine the count of one-to-one matches, we only consider region pairs with an IOU score above a defined threshold of 0.5. Consequently, we calculate the Detection Rate (DR), the Recognition Accuracy (RA), and the F-measure (FM) determined by combining DR and RA, following~\cref{eq:detection}.
\begin{equation}
        \mathcal{FM} = \frac{2.\mathcal{DR}.\mathcal{RA}}{\mathcal{DR}+\mathcal{RA}}; \quad \text{with} \quad \mathcal{DR} = \frac{o2o}{N}; \quad \text{and} \quad \mathcal{RA} = \frac{o2o}{M}
  \label{eq:detection}
\end{equation}
where, $o2o$ is the number of counting one-to-one matched pairs, $N$ is the number of boxes in the ground truth, $M$ is the number of bounding boxes detected by each baseline model.
For the performance of pre-trained YOLO models YOLOv${5}$, YOLOv${8}$, YOLOv$_{10}$ are evaluated using common metrics such as precision, recall, mAP50, and mAP50-95. mAP50 measures the overlap between predicted and actual bounding boxes, considering a match correct if the overlap 50\% or more. Meanwhile, mAP50-95 provides a more detailed accuracy assessment across varying overlap thresholds from 50\% to 95\%.

\noindent\textbf{Baseline Models.}
First, the enhanced YOLOv$_{1}$ architecture is a single neural network designed to predict object class probabilities and bounding boxes simultaneously. It processes input images of 1472x1472 pixels with 3 channels and outputs a grid containing class probabilities and bounding boxes for each cell. The model, inspired by~\cite{redmon2016you}, is built as a convolutional neural network (CNN) and evaluated using the KhmerST dataset. It applies a series of convolutional layers, followed by LeakyReLU activation and max-pooling for down-sampling, gradually increasing the number of filters. The input images have a fixed size of 1472x1472 pixels with 3 channels, and the model produces a tensor of predictions with a size of 9x23x23 as the final output. The 23x23 represents the number of output grids, and 9 represents 1 for the existence of the object and 8 values for the polygon coordinates of $x$ and $y$.

\begin{figure}[H]
  \centering
  \begin{subfigure}{.94\textwidth}  
    \includegraphics[width=\linewidth]{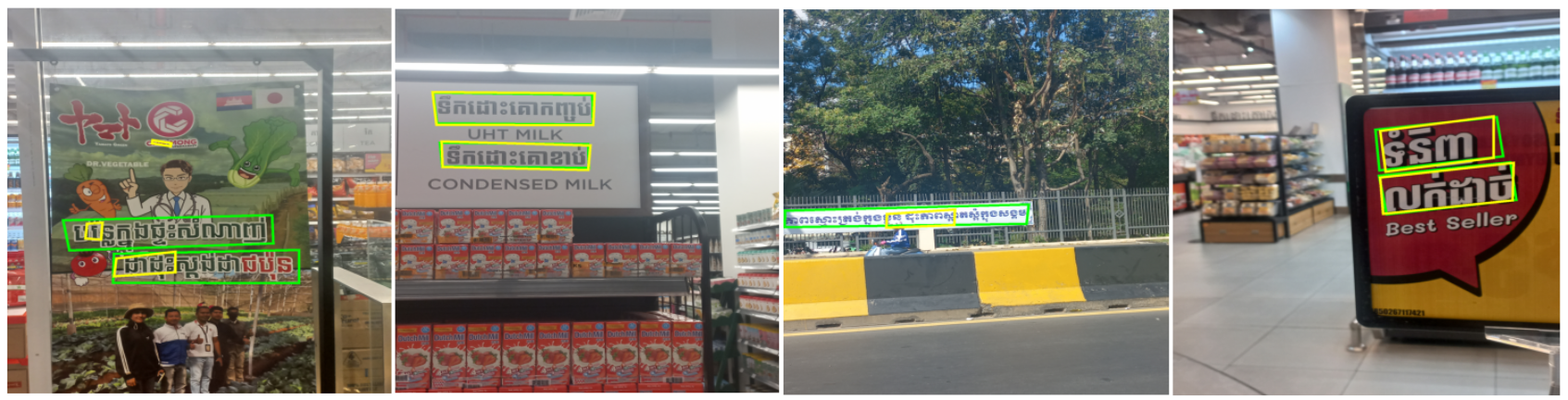}
    \caption{Detection results from YOLOv$_{1}$ enhanced with CNN: the model predicts the text area in each image by drawing bounding boxes.}
    \label{fig:short-a}
  \end{subfigure}
  
  \begin{subfigure}{.94\textwidth}  
    \includegraphics[width=\linewidth]{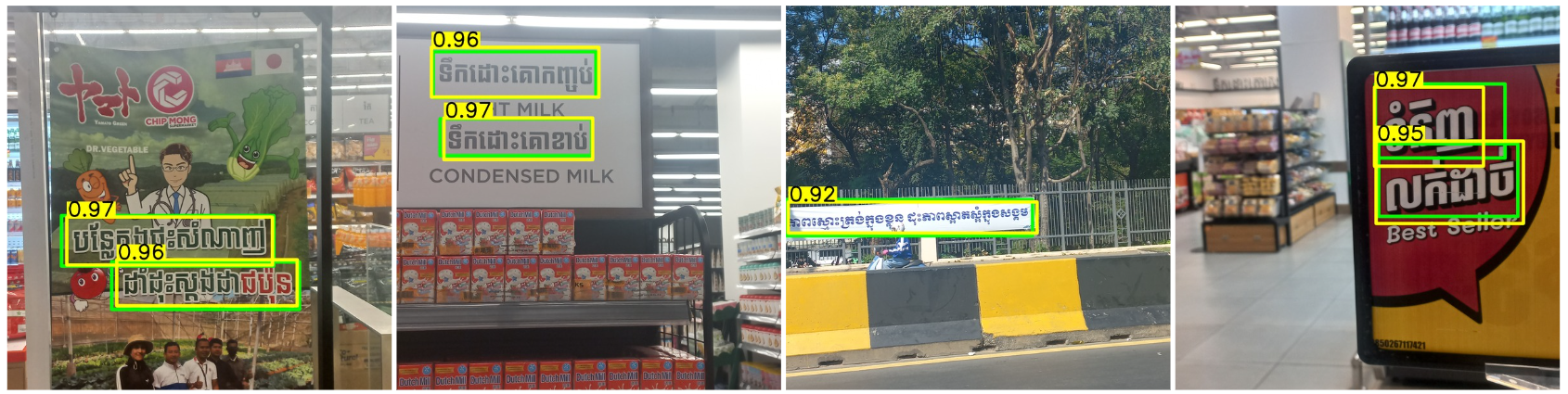}
    \caption{In the detection output of the YOLOv$_{5}$ model, we detect the text areas by drawing the bounding boxes with mAP values.}
    \label{fig:short-b}
  \end{subfigure}
  
  \begin{subfigure}{.94\textwidth}  
    \includegraphics[width=\linewidth]{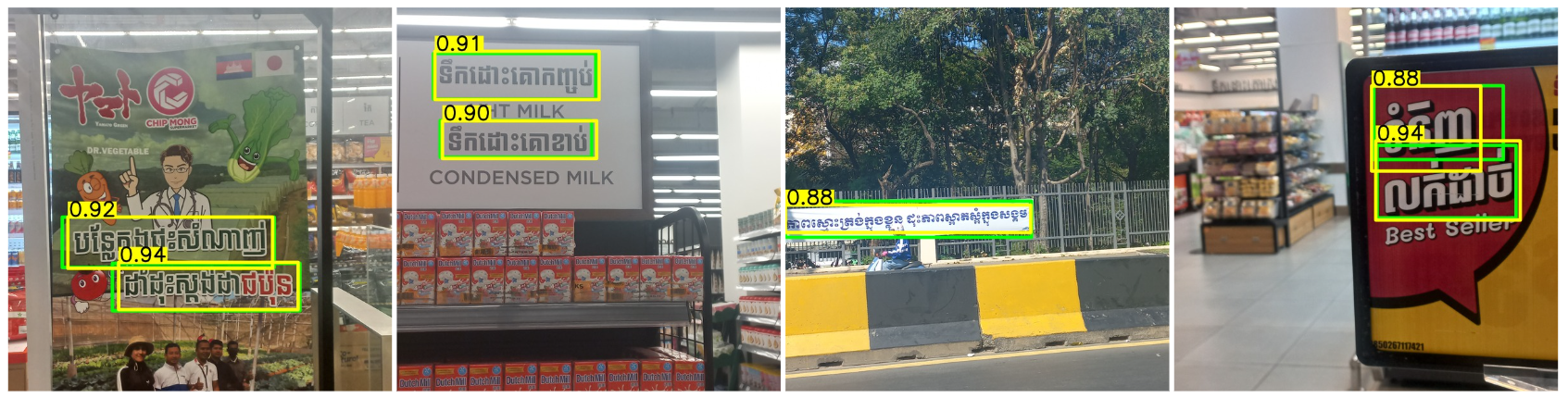}
    \caption{In the detection output of the YOLOv$_{8}$ model, we detect the text areas by drawing the bounding boxes with mAP values.}
    \label{fig:short-c}
  \end{subfigure}

  \begin{subfigure}{.94\textwidth}  
    \includegraphics[width=\linewidth]{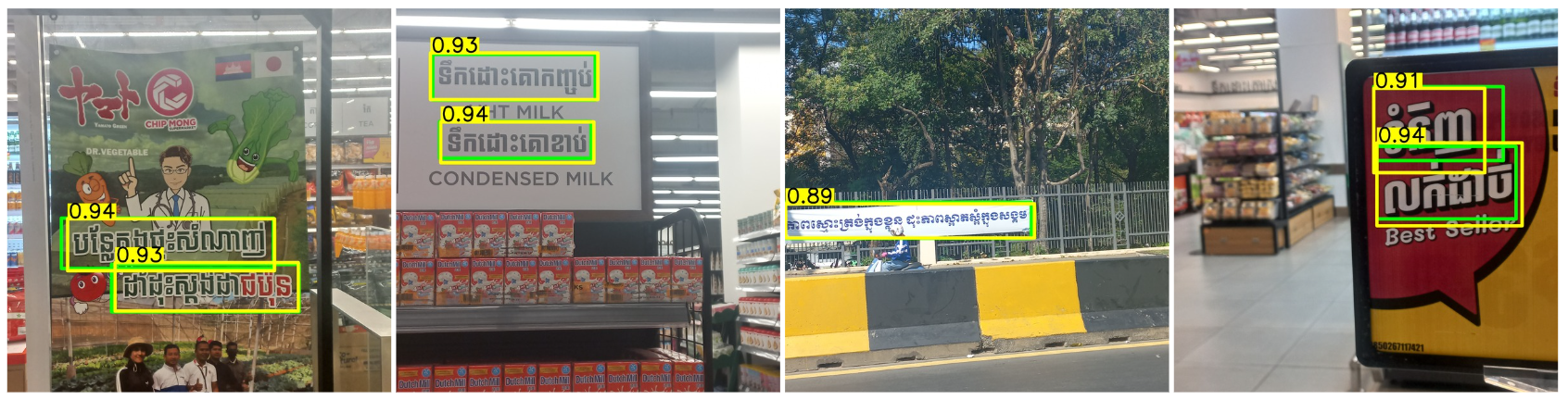}
    \caption{In the detection output of the YOLOv$_{10}$ model, we detect the text areas by drawing the bounding boxes with mAP values.}
    \label{fig:short-d}
  \end{subfigure}
  \caption{The output of the YOLO models: For each image, YOLOv$_{1}$ predicts the text areas, and pre-trained models like YOLOv$_{5}$, YOLOv$_{8}$, and YOLOv$_{10}$ also produce mAP values. The images with green bounding boxes represent the ground truth, while those with yellow bounding boxes show the predictions from the model.}
  \label{fig:6}
\end{figure}

We experimented with the model as a modular list with different combinations of layer numbers (1, 2, 3) and filter amounts (8, 16, 24) randomly. As shown in \cref{table1}, the reported results indicate that 2 layers with 24 filters produce the best detection rate of 0.733, a recognition accuracy of 0.866, and an F-Measure score of 0.794 compared to other variations of the model. The results of text-line detection using the enhanced YOLOv$_{1}$ with a CNN architecture are displayed in \cref{fig:short-a}. We observe that the bounding boxes are correctly positioned over the text areas, though in some cases, the model didn't predict the correct text area.

\begin{table}[tb]
  \caption{The performance of the enhanced YOLOv$_{1}$+CNN architecture.}
  \label{table1}
  \centering
  \begin{tabular}{lcccc}
    \toprule
    Nb. Layers & Nb. Filters & DR & RA & FM \\
    \midrule
        1 & 8 & 0.694 & 0.819 & 0.751 \\
        2 & 24 & 0.733 & 0.866 & 0.794 \\
        3 & 24 & 0.627 & 0.828 & 0.714 \\
  \bottomrule
  \end{tabular}%
\end{table}

\begin{table}[tb]
  \caption{Performance comparison on different versions of YOLO models.}
  \label{table2}
  \centering
  \begin{tabular}{lcccccc}
    \toprule
    Model & Precision & Recall & mAP50 & mAP50-95 & Runtime(H) & Params(M)\\
    \midrule
        YOLOv$_{5}$ & 0.847 & 0.787 & 0.875 & 0.591 & 0.56  & 7\\
        YOLOv$_{8}$ & 0.873 & 0.832 & 0.899 & 0.625 & 1.293  & 11\\
        YOLOv$_{10}$ & 0.905 & 0.76 & 0.87 & 0.593  & 1.965 & 8\\
  \bottomrule
  \end{tabular}%
\end{table}


Second, we fine-tuned the pre-trained YOLOv${5}$, YOLOv${8}$, and YOLOv$_{10}$ on the KhmerST dataset. Each image was resized to 640x640 pixels, and the corresponding JSON files were pre-processed into a text file format containing the image filename and coordinates, including the class ID, $x\_center$, $y\_center$, $width$, and $height$, where class ID "0" represents Khmer text as an object. As shown in~\cref{table2}, YOLOv${10}$ achieved the highest precision value of 0.905, while YOLOv${8}$ excelled in recall, mAP50, and mAP50-90, with values of 0.832, 0.899, and 0.865, respectively. However, YOLOv${10}$ recorded the lowest recall score of 0.760, compared to YOLOv${5}$, which had a recall rate of 0.787, and 0.832 of YOLOv${8}$. Overall, while YOLOv${5}$ offers efficiency with balanced performance, YOLOv${8}$ and YOLOv${10}$ provide progressively more complex architectures and higher precision, albeit with increased computational demands and runtime. The Fine-tuned YOLO models predict the bounding region of each image by providing the bounding box values along with the mAP score. The results of detection using the fine-tuned versions of YOLOv$_{5}$ are shown in \cref{fig:short-b}, YOLOv$_{8}$ in \cref{fig:short-c}, and YOLOv$_{10}$ in \cref{fig:short-d}.

Overall, we observed that the best model among the YOLO versions is YOLOv${8}$. This model achieved the highest recall, around 0.832, and a mean average precision (mAP) of 0.899, although its precision of 0.873 was slightly lower than that of YOLOv${10}$, which had a precision of 0.905. However, YOLOv${8}$ outperforms YOLOv${10}$ in text detection due to several key factors. Firstly, YOLOv${8}$ excels at detecting small objects~\cite{hussain2024yolov5}, which is critical for text detection, as text often appears as small elements within images. Additionally, YOLOv${8}$'s 11M parameters allow it to capture more detailed features and model complexities, which are essential for accurately recognizing text with various fonts, sizes, and orientations. While YOLOv${5}$, with its 7M parameters, and YOLOv${10}$, with 8M parameters, may offer reduced latency, which comes at the cost of accuracy in text detection. 
In contrast, YOLOv${8}$'s ability to handle the intricacies of small object detection makes it more suitable for this task, even with slightly higher latency, while it is also more efficient in terms of computational resources and time. 
Moreover, YOLOv${1}$ also demonstrated a relatively good ability to detect text areas in images, with some limitations when dealing with images of low resolution and complex backgrounds, as illustrated in \cref{fig:short-a}.

\begin{table}[t]
  \caption{Character Error Rate (CER) and Word Error Rate (WER) performances for TrOCR and Tesseract OCR.}
  \label{table3}
  \centering
  \begin{tabular}{lcc}
    \toprule
     Model & CER (\%) & WER (\%)\\
    \midrule
    TrOCR & 1.01 & 2.24\\
    Tesseract & 1.30 & 4.75\\
    \bottomrule
  \end{tabular}
\end{table}

\subsection{Scene Text Recognition}
Scene Text recognition involves the identification and extraction of textual information. In this matter, we intend to investigate the performance of two different methodologies on our proposed KhmerST dataset. 

\noindent\textbf{Evaluation Metrics.}
For the recognition phase, we calculate the character error rate (CER) and word error rate (WER) as the ratio of unrecognized characters over the total number of characters in ground truth. It typically includes insertions, deletions, and substitutions. CER and WER is calculated as follows:
\begin{align}
    \displaystyle
        \text{CER} = \frac{\text{S$_{c}$}+\text{D$_{c}$}+\text{I$_{c}$}}{\text{N$_{c}$}}; \quad
        \text{WER} = \frac{\text{S$_{w}$}+\text{D$_{w}$}+\text{I$_{w}$}}{\text{N$_{w}$}}
 \label{eq:CER_WER}
\end{align}
where $S$ The number of substitutions $D$ is the number of deletions $I$: The number of insertions $N$ The total number of characters in ground truth. 

\noindent\textbf{Baseline Methods.}
First, we experimented with the TrOCR pre-trained model, which consists of a Transformer-based encoder and an auto-regressive text Transformer decoder to perform optical character recognition (OCR). The TrOCR model is designed to understand the context and structure of written language. It has shown its superior performance compared to the current state-of-the-art models in printed, handwritten, and scene text recognition tasks~\cite{li2023trocr}. TrOCR achieved a relatively good performance with an overall CER of 0.90 and WER of 1.02 as shown in \cref{table3}. We calculated the CER and WER using \cref{eq:CER_WER}. Second, we utilize the Tesseract tool, using its OCR capabilities to extract text from natural scene images. With the KhmerST dataset, Tesseract is able to achieve a CER of 1.30 and WER of 4.75. The results denoted in \cref{table4} show the outputs produced by the TrOCR pre-trained model and Tesseract. We randomly selected five images from the test set to demonstrate that the models were able to extract the text correctly in some cases, but failed to do so in others.

Overall, we observed that state-of-the-art models and tools such as the TrOCR pre-trained model and Tesseract tool, did not perform well with our KhmerST dataset. Both TrOCR and Tesseract faced significant difficulties in extracting Khmer scene text accurately. These challenges can be attributed to the unique characteristics of the Khmer script, including its complex ligatures, varying baseline, and intricate diacritics. Additionally, the presence of diverse backgrounds and varying text orientations in natural scenes further the recognition difficulties. 
Our observations highlight the need for more specialized OCR models that can handle the complexity of the Khmer script, including the fonts and their appearances, as well as the diversity of scene images.


\begin{table}[t]
  \caption{Examples of text recognition using the TrOCR model and Tesseract compared with the ground truth. Errors in the predictions are highlighted in red.}
  \label{table4}
  \centering
  \begin{tabularx}{\textwidth}{@{}lXXXX@{}}
    \toprule
    Instance & \quad \quad \quad \quad Ground-Truth & \quad \quad \quad \quad Tesseract & \quad \quad \quad TrOCR \\
    \midrule
    \multicolumn{4}{c}{\includegraphics[width=\textwidth]{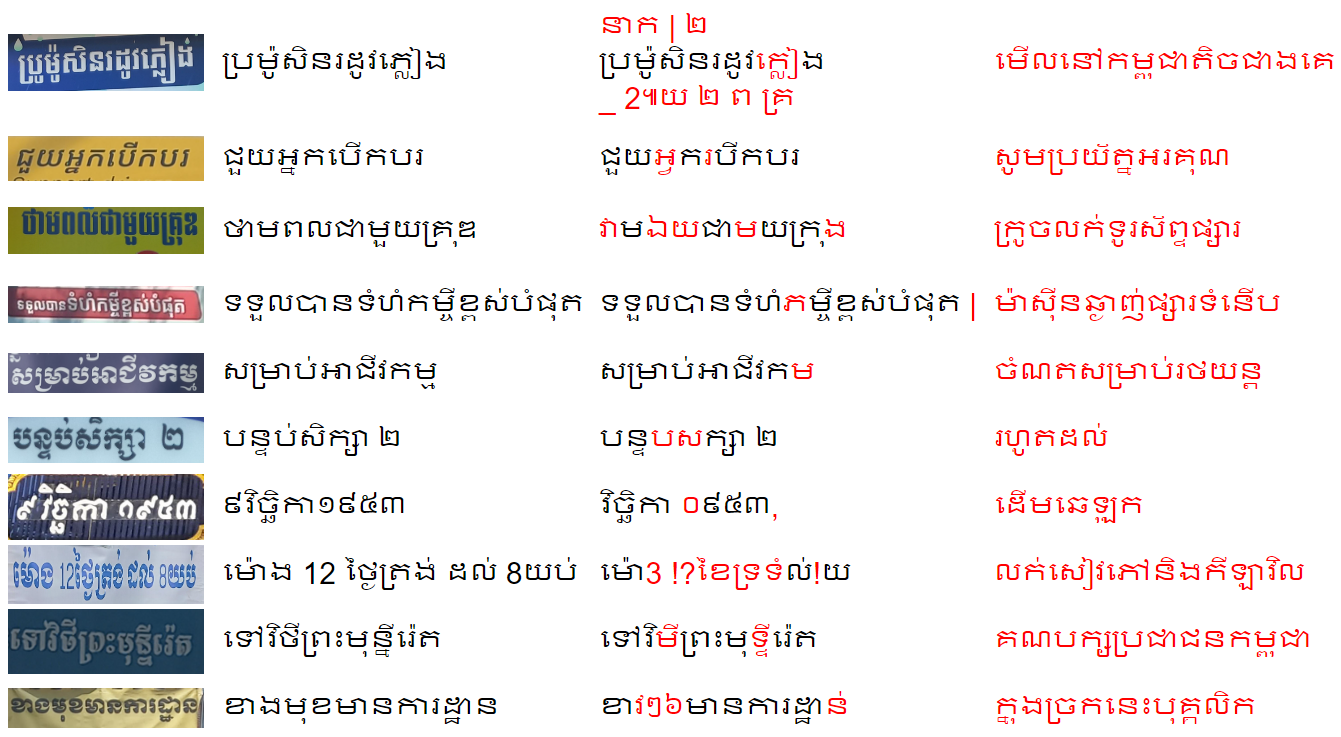}} \\
    \midrule
  \end{tabularx}
\end{table}

\subsection{Limitations}
\label{sec:limitations}
The pre-trained TrOCR model, with its approximately 159 M parameters, requires a substantial amount of data for effective training and testing. Given the limited size of the KhmerST dataset, training the TrOCR model led to over-fitting. Over-fitting occurs because the model memorizes the training data rather than learning to generalize from it, resulting in poor performance on unseen data. Additionally, the large model size contributes to increased computational demands, making it less feasible for environments with limited resources. Conversely, Tesseract OCR, an open-source optical character recognition engine, performs well with well-resourced languages such as English~\cite{mahajan2022natural, saoji2021text}, Hindi~\cite{shwait2022detection}, and Chinese~\cite{yun2021yolov3}. These languages benefit from extensive annotated datasets and well-developed linguistic models, enabling Tesseract OCR to accurately extract text. However, for the low-resource Khmer language, Tesseract OCR's performance is sub-optimal. As shown in \cref{table4}, the outputs from both the TrOCR model and Tesseract OCR did not align with the ground truth text for Khmer.

Several factors contribute to the poor performance of these state-of-the-art models with Khmer text in natural scene images. Firstly, the Khmer script is inherently complex, with intricate characters, multiple diacritics, and unique word formation rules. Unlike Latin-based scripts, Khmer characters combine in various ways, creating numerous glyphs that must be recognized, as shown in \cref{fig:1}. In addition, Khmer text in natural scenes often appears in diverse font styles, sizes, and orientations. This variability significantly challenges OCR models, as they need to adapt to different typographic representations of the same characters. Another issue is the lack of standardization in Khmer text, which can vary widely in terms of spelling, formatting, and orthographic conventions. This lack of standardization increases the difficulty of creating robust models that can handle all possible variations. Environmental factors also play a role: text in natural scenes is subject to various distortions due to lighting conditions, shadows, reflections, and occlusions. These factors can obscure parts of the text, making it harder for OCR models to accurately detect and recognize the text.

\section{Conclusion}
\label{sec:conclusion}

This research work introduces the KhmerST Dataset, the first scene text dataset for the Khmer language, which contains around 1,544 images. The dataset is divided into two main categories: indoor 997 images and outdoor 547 images. We annotated the text in the scene images at the line-level and stored the coordinates as polygons. The dataset was collected by capturing real images with various fonts, text sizes, and backgrounds. These present a significant challenge for text detection and recognition systems. We believe that our dataset will be a great resource for improving OCR and advancing research in Khmer STDR. In addition to the dataset, we also produced a text detection and recognition benchmark and discussed the performance limitations of the current state-of-the-art models on the KhmerST dataset. To deal with such a challenging dataset, STDR require different models that can handle the unique characteristics of the Khmer script and the diverse conditions in which the text appears.

Addressing the challenges of Khmer text recognition in natural scene images requires a multifaceted approach. Future research will focus on: (i) expanding the dataset with more diverse images and text styles to provide a comprehensive resource for the research community. Additionally, generating synthetic data will augment the existing KhmerST dataset, overcoming its size limitations; (ii) developing specialized architectures tailored to the intricacies of the Khmer script aims to enhance detection and recognition accuracy by incorporating specific linguistic and typographic features; (iii) emphasizing multimodal approaches that integrate visual and textual data will further refine text recognition capabilities, particularly in disambiguating complex text scenarios. These areas for future contributions are pivotal for advancing Khmer STDR in natural images, ensuring that OCR models effectively manage the intricacies and variations of this low-resource language.

\begin{credits}
\subsubsection{\ackname} This research study is supported by the France Government Scholarship, co-funded by the Cambodia Academy of Digital Technology (CADT).
We would like to thank La Rochelle University, Laboratoire Informatique Image Interaction (L3i), for their funding in annotating the dataset.
\end{credits}

\clearpage
\bibliographystyle{plain}
\bibliography{references}

\end{document}